\documentclass[10pt,twocolumn,letterpaper]{article}

\usepackage{wacv}
\usepackage{times}
\usepackage{epsfig}
\usepackage{graphicx}
\usepackage{amsmath}
\usepackage{amssymb}
\usepackage{booktabs}
\usepackage[accsupp]{axessibility}
\def\wacvPaperID{1937} % Enter the WACV Paper ID here

\wacvalgorithmstrack   % Uncomment this line if you are submitting to the Algorithms Track.
\wacvfinalcopy % *** Uncomment this line for the final submission

\ifwacvfinal
\usepackage[breaklinks=true,bookmarks=false]{hyperref}
\else
\usepackage[pagebackref=true,breaklinks=true,colorlinks,bookmarks=false]{hyperref}
\fi

\newcommand{\myparagraph}[1]{\vspace{2pt}\noindent{\bf #1}}

\begin{document}

%%%%%%%%% TITLE
\title{Learning Attention Propagation for Compositional Zero-Shot Learning}

% \and

\author{\vspace{0em}
\setlength\tabcolsep{0.1em}
% \pagenumbering{gobble}
\begin{tabular}{cccccc} 
Muhammad Gul Zain Ali Khan$^{1,3,4} $ & Muhammad Ferjad Naeem$^{2}$ & Luc Van Gool$^2$ & \\ A. Pagani$^{1, 3}$ & Didier Stricker$^{1,3,4}$ & Muhammad Zeshan Afzal$^{1,3,4}$\tabularnewline
\end{tabular}
\\
\renewcommand{\arraystretch}{0.5}
\begin{tabular}{cccc} 
    $^1$\normalsize{DFKI,} & $^2$\normalsize{ETH Zürich,} & $^3$\normalsize{TU Kaiserslautern,} &$^4$\normalsize{MindGarage}  %\tabularnewline
    %\normalsize{Yonsei University} & \normalsize{\enskip LINE Plus Corp.} & \normalsize{\enskip NAVER Corp.} & \normalsize{}
\end{tabular}
}%

\maketitle
\thispagestyle{empty}

%%%%%%%%% ABSTRACT
\begin{abstract}
   Compositional zero-shot learning aims to recognize unseen compositions of seen visual primitives of object classes and their states. While all primitives (states and objects) are observable during training in some combination, their complex interaction makes this task especially hard. For example, wet changes the visual appearance of a dog very differently from a bicycle. Furthermore, we argue that relationships between compositions go beyond shared states or objects. A cluttered office can contain a busy table; even though these compositions don't share a state or object, the presence of a busy table can guide the presence of a cluttered office. We propose a novel method called Compositional Attention Propagated Embedding (CAPE) as a solution. The key intuition to our method is that a rich dependency structure exists between compositions arising from complex interactions of primitives in addition to other dependencies between compositions. CAPE learns to identify this structure and propagates knowledge between them to learn class embedding for all seen and unseen compositions. In the challenging generalized compositional zero-shot setting, we show that our method outperforms previous baselines to set a new state-of-the-art on three publicly available benchmarks.
\end{abstract}

%%%%%%%%% BODY TEXT
\section{Introduction}
Dog species differ considerably from each other. However, when presented with an unseen dog specie, we humans can recognize its states without hesitation. A child that has seen a wet car can recognize a wet dog regardless of the vast difference in appearance. Humans excel at recognizing previously unseen compositions of states and objects. This remarkable ability arises from our ability to reason about various aspects of objects and then generalize them over previously unseen objects. In zero-shot learning, the goal is to predict unseen classes, having seen a set of seen classes and a description of all the classes. A vector of attributes for all classes is provided in the most common configuration of zero-shot learning. The task is to learn the mapping between class description vectors and images such that it can be generalized over unseen classes~\cite{sorbello1970annhuman_firstbook,shanmuganathan2016annhuman_book,noriega2005mannhuman_mlp}. Although deep neural networks have been modelled after the human mind~\cite{sorbello1970annhuman_firstbook,shanmuganathan2016annhuman_book,noriega2005mannhuman_mlp}, they struggle to perform well in zero-shot learning. In this paper, we study a special setting of zero-shot learning called Compositional Zero-Shot Learning.\\

\begin{figure}[t]
\centering
\includegraphics[scale=0.85]{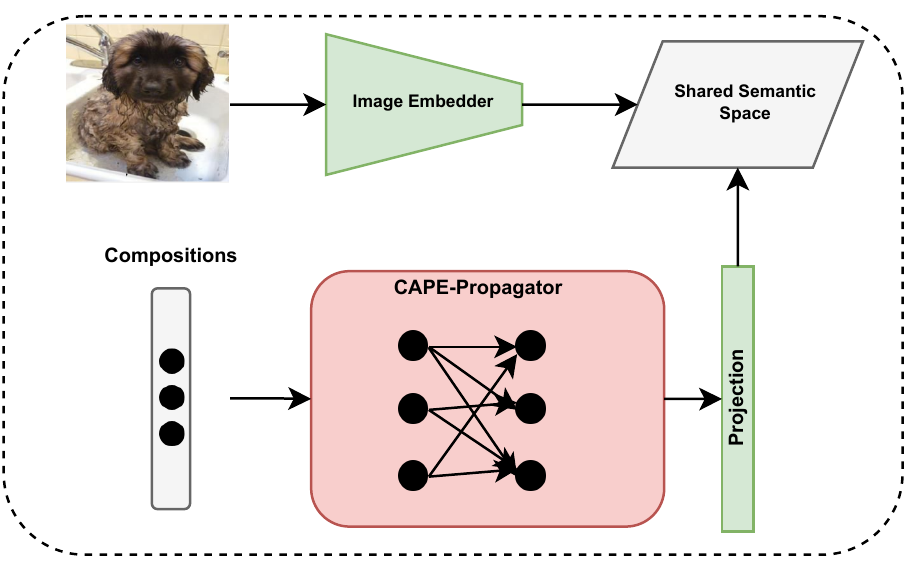}
\caption{Shows overview of our approach. CAPE-Propagator exploits self-attention mechanism to learn interdependency structure between compositions by identifying critical propagation routes. We project output of self attention into a shared semantic space along with image embedding. Compositions that are similar to each other are placed near and far away from other compositions in shared semantic space.}
\label{fig:overview_diagram}
\end{figure}

``Compositional" in Compositional Zero-Shot Learning derives from the composition of primitives of objects and their states. At training time, all primitives (state, object) are provided in some combination but not all compositions. The goal is to predict novel compositions of primitives during test time. This poses several challenges because of the complex interaction between objects and their possible states. For example, a wet car is very different from a wet dog in visual appearance. Furthermore, a composition can be abstract as well, i.e., Old city, Ancient Tower. In real-world settings, multiple valid compositions can be found in one image, i.e. A clean desk in an image of a wet dog. A successful methodology should be able to learn all aspects of the complex interaction of objects and their states. \\
One simple solution to the above mentioned challenges is to disentangle states from objects. If states can be completely disentangled from objects, new combinations of states and objects can be predicted easily. This approache have been studied in some recent works~\cite{atzmon2020causal,ruis2021independentprop} on synthetic or simple datasets like UT-Zappos~\cite{yu2014utzap1,yu2017utzap2}. In recent work, contrastive loss function-based methodology is introduced~\cite{li2022scen} for real-world datasets. The approach in \cite{li2022scen} uses a generative network to generate novel samples in order to bridge the domain gap between seen and unseen samples. In the real world, states can be imagined as transformation functions of objects. A dry car changes the visual appearance after the application of state wet. Approaches in \cite{nagarajan2018attributesasoperators,li2020symnet} have studied the function of states as a transformation function. Other methodologies have tried to exploit all available information to learn useful correlations~\cite{purushwalkam2019tmn,misra2017le+}. Recent works have explored learning dependency structures between word embeddings and propagating them to unseen classes~\cite{naeem2021cge,mancini2021compcos}.\\
% now we talk about ourselves
In the real world, compositions do not occur in isolation. They are intricately entangled with each other and full of noise. No approach has considered the holistic view of complex interactions between compositions or their primitives. While the approach in \cite{naeem2021cge} exploits the shared information between compositions that share primitives, it still ignores the interaction of compositions that do not share primitives, such as a coiled plate can be found in a cluttered kitchen or a narrow road can be seen in an ancient city. We study a more holistic view of interactions between compositions. We argue that there is a hidden interdependency structure between compositions. An overview of our approach is shown in Figure. \ref{fig:overview_diagram}. We exploit self-attention mechanism to explore hidden interdependency structure between primitives of compositions and propagate knowledge between them. Our contributions are as follows: 
\begin{itemize}
    \item We propose a multi-modal approach that learns to embed related compositions closer and unrelated far away. 
    \item We propose a methodology that learns the hidden interdependency structure between compositions by learning critical propagation routes and propagates knowledge between compositions through these routes. 
    \item Unlike \cite{naeem2021cge}, our approach does not require prior knowledge of how the compositions are related. 
\end{itemize}

\section{Related Work}
Recent works have exploited the fundamental nature of states and objects to build novel algorithms. This includes reasoning over the effect of states over objects~\cite{li2020symnet,nagarajan2018attributesasoperators}. The approach introduced in \cite{nagarajan2018attributesasoperators} considers states as a linear function that transforms objects into compositions. These linear functions can add a state to a composition or remove a state from the composition by inverting the linear function. Approach in \cite{li2020symnet} also considers the symmetry of states. Both approaches~\cite{nagarajan2018attributesasoperators,li2020symnet} exploit group theory principles like closure, associativity, commutativity and invertibility. Both approaches~\cite{nagarajan2018attributesasoperators,li2020symnet} use triplet loss~\cite{hoffer2015tripletloss} as the objective function. In contrast with \cite{nagarajan2018attributesasoperators}, \cite{li2020symnet} uses a coupling and decoupling network to add or remove a state from a composition. Other approaches have tried to exploit the relationship between states and objects instead of assuming states as transformative functions~\cite{atzmon2020causal,ruis2021independentprop,kipf2016gcn,mancini2021compcos,misra2017le+,purushwalkam2019tmn}.\\
The approach in \cite{misra2017le+} argues that context is essential to model, i.e. red in red wine is different from red in red tomato. The approach in \cite{misra2017le+} argues that compositional classifiers lie in a smooth plane where they can be modelled and propose to model compositional classifiers of primitives using SVMs. These classifiers are pre-trained using SVMs and then fed into a transformation network that translates them into compositional space. The transformation network is three layered non-linear Multi-Layer Perceptron (MLP). Final predictions are retrieved by a simple dot product between the output of the transformation network and image embedding. One recent work uses word embeddings of primitives, and a simple MLP projects embeddings into a shared semantic space~\cite{mancini2021compcos} (Compcos). Compcos~\cite{mancini2021compcos} proposes to replace logits with cosine similarity between image embedding and projected word embeddings. Another recent work proposes to replace multi-layer perceptron with a Graph Convolutional Network~\cite{kipf2016gcn} (GCN) to model the interaction between compositions (CGE). CGE~\cite{naeem2021cge,mancini2022gcnopen} argues that compositions are entangled by their shared primitives and proposes to use GCN that propagates knowledge through entangled compositions. CGE~\cite{naeem2021cge} utilizes a dot product based compatibility function between compositional nodes and image embeddings to calculate the scores of compositions. While CGE~\cite{naeem2021cge} uses average of state and object embeddings as compositional embeddings, Compcos~\cite{mancini2021compcos} uses concatenation of state and object embeddings to represent a composition. \\
Another view for solving CZSL problem is to disentangle states from objects~\cite{atzmon2020causal,ruis2021independentprop,li2022scen}. Approach in \cite{atzmon2020causal} proposes to exploit causality~\cite{scholkopf2022causalforml,zhang2020causalann,yang2021causalvae,bengio2019causalmeta,goudet2018causalbook,pawlowski2020causalstructural,parascandolo2018causallearningindep} to disentangle states and objects. The causal view in \cite{atzmon2020causal} assumes that compositions are not the effect of images but rather the cause of images, and do-intervention on primitives generates a distribution from which a given image can be sampled. Another recent work proposes to learn independent prototypes of states and objects by enforcing independence between the representation of states and objects~\cite{ruis2021independentprop}. This approach~\cite{ruis2021independentprop} further exploits a GCN to propagate prototypes of states, objects and their calculated composition. Like CGE~\cite{naeem2021cge}, the approach in ~\cite{ruis2021independentprop} calculates scores of compositions by a dot product between compositional nodes of GCN and image embeddings. This leads to further knowledge sharing between completely independent prototypes. Approaches in ~\cite{atzmon2020causal,ruis2021independentprop} focuses on synthetic dataset like Ao-Clevr~\cite{atzmon2020causal} or UT-Zappos~\cite{yu2014utzap1,yu2017utzap2}. A recent approach has explored the disentanglement of states and objects on real-world datasets (SCEN)~\cite{li2022scen}. SCEN~\cite{li2022scen} uses contrastive loss and proposes to use compositions with shared primitives as positive samples and others as negative samples. SCEN~\cite{li2022scen} further proposes using a generative network to create novel compositions to bridge the gap between seen and unseen compositions. Our methodology is closer to CGE~\cite{naeem2021cge} that proposes to model the interdependency between compositions. However, CGE~\cite{naeem2021cge} only considers a dependency based on shared primitives. CGE~\cite{naeem2021cge} does not model more complex interdependencies that do not share primitives. Such as, the presence of a cluttered desk can guide the presence of cluttered office or the presence of a coiled plate can guide the presence of cluttered kitchen. The major limitation of CGE~\cite{naeem2021cge} is the usage of GCN to model interdependency structure because GCN relies on a fixed adjacency matrix to hardcode the interdependency structure. We propose to exploit self-attention mechanism like the one proposed in Transformer network~\cite{vaswani2017attentionisallyouneed} to learn this interdependency structure instead.\\
% \subsection{Transformers}
Transformer networks were first introduced for natural language processing to solve the problem of forgetting past hidden states for long sentences and vanishing gradient posed by RNNs~\cite{vaswani2017attentionisallyouneed}. Transformer networks~\cite{vaswani2017attentionisallyouneed} use a stack of encoder and decoder blocks that contain Multi-Head attention and multi-layer perceptrons. Multi-Head Attention (MHA) calculates attention on slices of features from query, key and value pairs. Number of slices are determined by the number of heads in MHAs. The usage of transformers~\cite{vaswani2017attentionisallyouneed} in the image classification task was explored in ~\cite{dosovitskiy2020imagetransformer} that proposed to divide an image into $16\times 16$ tokens. An encoder network extracts features from $16\times16$ tokens that are used for classification. This has lead to a number of methodologies that utilize transformer networks for image and video processing~\cite{radford2021clip,tu2022transformer1,zhang2022transformer2,liang2021transformer3,lee2021transformer4,wang2022transformer5,liu2021transformer6}. Large networks like proposed in \cite{radford2021clip} also have zero shot capabilities because they are trained on large datasets.\\

Our proposed approach builds on the findings of several prior works~\cite{misra2017le+,naeem2021cge,mancini2021compcos}. These works have explored the interdependency between compositions~\cite{naeem2021cge,misra2017le+,mancini2021compcos} in a simplistic manner. We propose a more holistic view of the interdependency structure between compositions. We argue that compositions are not simply related based on explicitly shared primitives. There is also implicitly hidden interdependency between compositions. We further propose a self-attention-based methodology to explore and exploit this interdependency structure between primitives of compositions and propagate the knowledge between them. Our approach learns this interdependency structure during training in an end-to-end manner and can find more various dependencies between compositions than simply shared primitives. 
\section{Approach}
\label{sec:approach}
\subsection{Problem Formulation}
Firstly, we formally define Compositional Zero Shot Learning. Let $x\in \mathcal{X}$ denote an RGB image and $y\in \mathcal{Y}$  denote compositional label $y=(s,o)$ where $s \in \mathcal{S}$ is state and $o \in \mathcal{O}$ is object. We can then define training set $\tau$ as $\tau = \{(x,y)|x\in \mathcal{X}, y\in \mathcal{Y}_s\}$ where $\mathcal{Y}_s \in \mathcal{Y}$ represents seen compositions. The task of CZSL is to predict novel compositions $\mathcal{Y}_n$ during test time having seen set of seen labels $\mathcal{Y}_s$. Novel compositions $\mathcal{Y}_n$ does not include any compositions from $\mathcal{Y}_s$ i.e, $\mathcal{Y}_n \cap \mathcal{Y}_s =\emptyset$. We consider a specialized case of this problem called generalized compositional zero shot learning that includes seen compositions as well during test time $\mathcal{Y} = \mathcal{Y}_n \cup \mathcal{Y}_s$.
\begin{figure}[htp]
\centering
\includegraphics[scale=0.6]{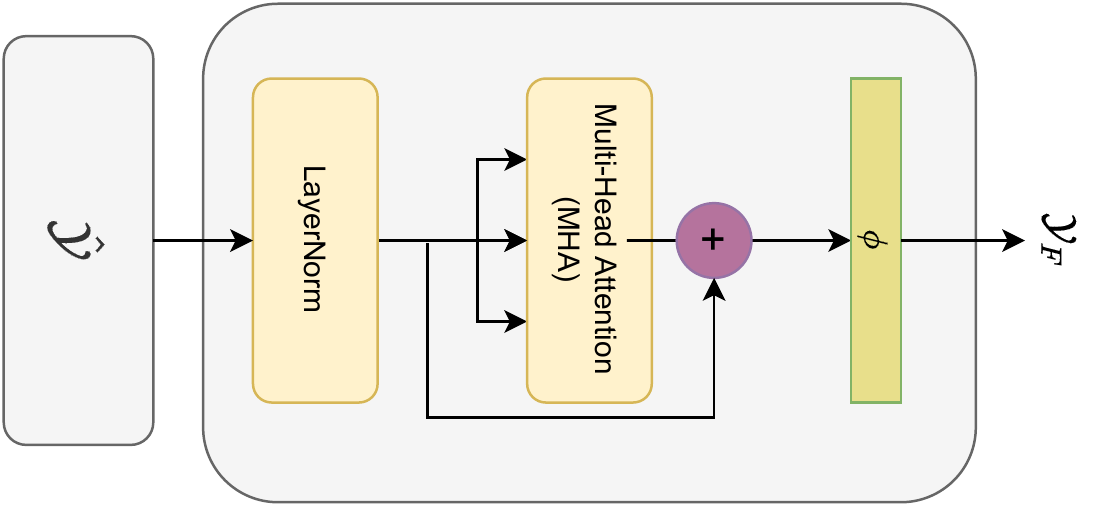}
\caption{The architecture of CAPE-Propagator module that finds propagation routes and projects embeddings into share semantic space. Embeddings are first passed through a LayerNorm~\cite{ba2016layernorm} followed by a Multi-Head Attention block. The output of Multi-Head Attention is residually added to the input and fed into a Multi-Layer Perceptron $\phi$ that projects it into a shared semantic space.}
\label{fig:cape-propagator}
\end{figure}
\subsection{Compositional Attention Propagated Embeddings (CAPE)}
\begin{figure*}[h]
    \centering
    \includegraphics[width=\textwidth]{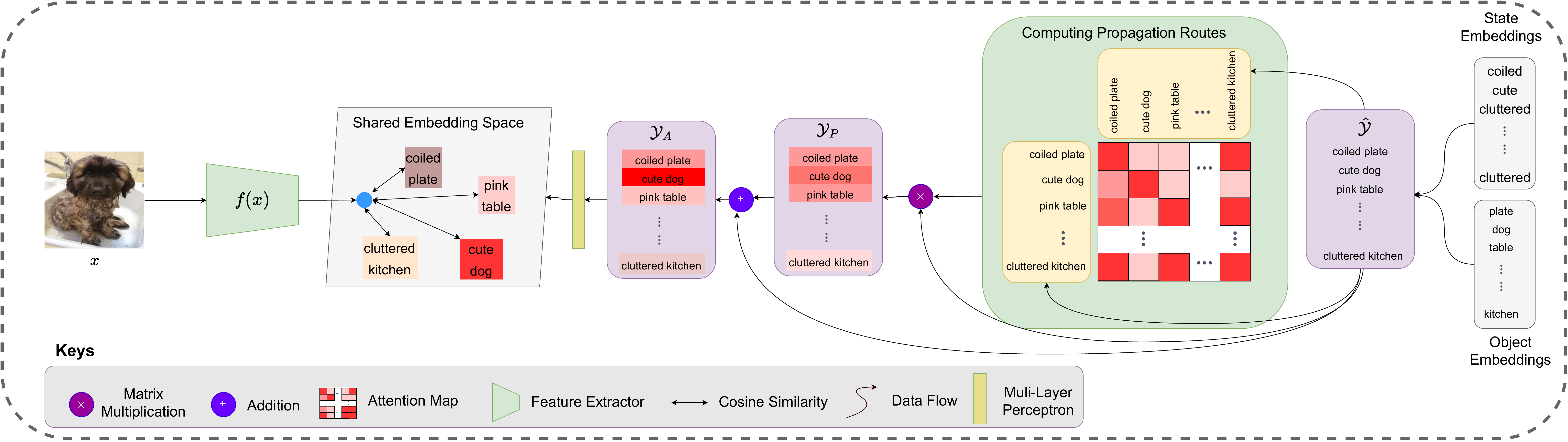}
    \caption{Compositional Attention Propagated Embedding (CAPE) learns interdependency between compositions. We concatenate word embeddings of states and objects to form embeddings of compositions $\mathcal{Y}$. During training, critical propagation routes are learnt by exploiting self-attention mechanism. These propagation routes are used to calculate an updated representation of embeddings in a residual manner. In the end, we learn to project these embeddings into shared semantic space along with image embeddings.}
    \label{fig:overall_architecture}
\end{figure*}

CZSL is the image classification task, where each image is associated with a composition of state ($s$) and object ($o$). 
In the most straightforward setting, the compositional primitives, i.e. states and objects, which are all observed during training, should provide an avenue for generalization. However, the interaction of states and objects is complex. For example, a visual transformation of dry to wet carpet is very different from a dry to the wet car. This means that our model needs to learn how each state and object interact with each other. Moreover, as noted in \cite{naeem2021cge}, compositions of states and objects are inherently a multi-label problem, e.g. while wet dog and dry car represent states with respect to liquid interaction, these objects simultaneously have other valid states representing colour, size etc. While not represented in the label set, these valid states, if discovered by the model, can present another avenue for knowledge transfer.\\
Furthermore, multiple compositions are a mixture of other compositions, e.g. An image of cluttered desk might also contain a blue mug. Learning a dependency structure between cluttered desk and blue mug will help propagate knowledge about blue mug to cluttered desk. Our novel model \textbf{Compositional Attention Propagated Embedding (CAPE)} aims to learn this knowledge propagation from training data as shown in Figure~\ref{fig:overall_architecture} leading to state-of-the-art performance. Unlike prior methods~\cite{naeem2021cge}, we do not limit the exploration of dependency structures by constraining our methodology by assumed priors (i.e. compositions are only related by shared primitives). We rely on our approach to discover all possible dependency structures during training.

\myparagraph{Learning the Propagation routes.} We introduce a module (CAPE-Propagator) for discovering interdependency structure between compositions. An overview of the CAPE-Propagator is given in Figure. \ref{fig:cape-propagator}. Given a list of compositional pairs, CAPE-Propagator is tasked with finding a propagation route to transfer knowledge between them before outputting the final embedding. We frame this as a self-attention problem. Given the set $\mathcal{Y}_s$, we find the compatibility between each composition as a query search problem. Each composition $\mathcal{Y}_{s_i}$ is defined in the feature space by a concatenation of the word embedding of its represented state $s$ and object $o$ as $\hat{\mathcal{Y}_{s_i}} \in \mathbb{R}^{ |Y| \times D}$ where $D$ is the feature dimension. Let $\mathcal{T}_{Q}$, $\mathcal{T}_{K}$ and $\mathcal{T}_{V}$ be linear transformations that maps from $D$ to $D$ as a transformation of input pre-trained word embedding. These transformations map the input to a new linear space suitable for propagation. We pass $\hat{\mathcal{Y_s}}$ through a LayerNorm layer followed by each of these transformation layers to get the Query $\mathcal{Q}$, Key $\mathcal{K}$ and Value $\mathcal{V}$.

For a given Compositional pair $y_i$,
we define a propagation route as a compatibility between its query $Q_i$ and all Keys $K$ as:
\begin{equation}
    \mathcal{P}_i = Softmax(\mathcal{Q}_i \cdot \mathcal{K}_j \text{ for } j \in |K|)
\end{equation}
$P_i$ is the propagation coefficient and defines the contribution of each composition to the output embedding of a given composition. In essence, we expect the model to learn that a wet dog is related to other wet animals and other properties of the contained state and object. At test time, the propagation coefficients for all Compositions $\mathcal{Y}$ are computed to form $\mathcal{P} \in \mathbb{R}^{ |\mathcal{Y}| \times |\mathcal{Y}|}$.

\myparagraph{Propagating the Compositional Knowledge.}  CAPE utilizes the Propagation coefficients $\mathcal{P}$ to propagate knowledge between compositions at test time. The propagated knowledge results in computing a new representation of each pair $\mathcal{Y}_P \in \mathbb{R}^{|\mathcal{Y}| \times D}$ as:
\begin{equation}
    \mathcal{Y}_P = \mathcal{P} \times \mathcal{V}
\end{equation}
This operation aggregates knowledge across all compositions defined in our dataset and makes the model aware of all the properties of each state and object. Since we learn the propagation coefficients from data, these propagation routes are improved during training on $\mathcal{Y}_{s}$ instead of having to hardcode them from our label set similar to previous methods\cite{naeem2021cge}. The propagated embeddings are added as a residual to the initial embedding to get the output Compositional Embedding $\mathcal{Y}_A = \hat{\mathcal{Y}} +  \mathcal{Y}_P$. This propagation is done for multiple heads where each head can learn to identify separate important properties between compositions. We set the number of heads to six for CAPE. In the end, a three-layer non-linear Multi-Layer Perceptron (MLP) projects concatenation of all the heads into a shared semantic space to get $\mathcal{Y}_F=\phi(\mathcal{Y}_A)$ where $\phi$ represents three layer non-linear MLP. For projection, $\phi$ expands the input embeddings $\mathcal{Y}_A$ into $4096$ dimensions, then projects it into the original dimension $D$. 
% This enables extrapolation of information and then extraction of useful information from it. Ferjad: Not scientific
Last layer of $\phi$ projects embeddings into shared semantic space and is defined as $\mathcal{W}_l\in \mathbb{R}^{ |D|} $ where $D = |f(x)|$ represents dimensionality of image features, $W_l$ represents weights of layer. Each of the first two layers are followed by LayerNorm~\cite{ba2016layernorm}, ReLu~\cite{agarap2018relu} and Dropout~\cite{srivastava2014dropout} where dropout rate is set at $p=0.5$. The last layer is followed by the activation function ReLu~\cite{agarap2018relu}.

\myparagraph{Measuring Compatibility of an Image to Composition.} Given an image $x$, we pass it through a learnable feature extractor $f$ to get feature respresentation $f(x)$. The compatibility of an Image to each composition is measured to get the score $s$:
\begin{equation}
    s(x, Y_i) = \frac{f(x) \cdot \mathcal{Y}_{F_i}}{|f(x)| |\mathcal{Y}_{F_i}|} 
    \label{eq:compatibiltiy}
\end{equation}

\myparagraph{Objective function.} We define a cross-entropy on top of our scoring function to learn the feature extractor $f$ and CAPE-Propagator in an end-to-end manner.
\begin{equation}
    L = -\log ( \frac{\exp{s(x, Y_i)}}
    {\sum_{j \in \mathcal{Y}_{s}} \exp{s(x, Y_j)}}  )
\end{equation}
By optimizing the full model end to end, CAPE learns to identify critical propagation paths between compositions leading to more generalizable embeddings. 

\section{Comparison with State of the Art}

\myparagraph{Datasets. }We evaluate our methodology on three standard benchmark datasets MIT-States~\cite{isola2015mitstates}, CGQA~\cite{naeem2021cge} and UT-Zappos~\cite{yu2014utzap1,yu2017utzap2}. MIT-States~\cite{isola2015mitstates} dataset was collected using an older search engine with limited human annotations and significant label noise. This dataset is more abstract than other datasets and includes intangible objects and states such as ancient city, old town. MIT-States~\cite{isola2015mitstates} consists of 53k images with 245 objects, 115 states and 1252 compositions. Out of all compositions, 300 compositions are in the validation set, and 400 compositions are in the test set. \\
The second dataset that we use is CGQA~\cite{naeem2021cge}. CGQA~\cite{naeem2021cge} is a relatively newer dataset with the largest composition space among all three datasets. It consists of 38k images, 453 states, and 870 objects. There are 6963 seen compositions, 1368 unseen compositions in the validation set and 1047 unseen compositions in the test set.\\
UT-Zappos~\cite{yu2014utzap1, yu2017utzap2} is relatively simpler dataset and contains 16 states and 12 objects. It has 29k images, 83 seen compositions, 15 unseen compositions in the validation set and 18 unseen compositions in the test set. 

% \subsection{Implementation Details}
\myparagraph{Implementation Details. }We use Resnet-18~\cite{he2016resnet} to extract 512 dimensional feature vector for each image. Resnet-18~\cite{he2016resnet} is pre-trained on ImageNet~\cite{russakovsky2015imagenet} dataset. We utilize word embeddings for states and objects. For UT-Zappos~\cite{yu2014utzap1,yu2017utzap2} and MIT-States~\cite{isola2015mitstates}, we use concatenation of FastText~\cite{bojanowski2017fasttext} and Word2Vec~\cite{mikolov2013word2vec1,mikolov2013word2vec2}. For CGQA~\cite{naeem2021cge}, we use only word2vec~\cite{mikolov2013word2vec1,mikolov2013word2vec2} embeddings to represent states and objects. We use PyTorch to implement our methodology. We use Adam Optimizer~\cite{kingma2014adamoptimzer} with initial learning rate of $5.0 \times 10^{-05}$ and batch size 30. We train all our models for 120 epochs. 

\myparagraph{Metrics. }We follow the evaluation setting proposed in ~\cite{purushwalkam2019tmn}. We evaluate our methodology on Area Under Curve (AUC), Harmonic Mean (HM), Seen Accuracy (S) and Unseen Accuracy (U). Seen accuracy is calculated on seen compositions, and unseen accuracy is calculated on unseen compositions. Harmonic Mean (HM) is calculated on Seen and Unseen accuracy. AUC is calculated based on the variation of calibration bias between seen and unseen compositions and represents performance at different operating points~\cite{li2022scen}.
\begin{table*}[h]
\centering
\begin{tabular}{|l|llll|llll|llll|}
\hline
 &
  \multicolumn{4}{c|}{\textbf{MIT-States}} &
  \multicolumn{4}{c|}{\textbf{UT-Zappos}} &
  \multicolumn{4}{c|}{\textbf{C-GQA}} \\ 
%   \hline
\textbf{Method} &
  \multicolumn{1}{l|}{\textbf{S}} &
  \multicolumn{1}{l|}{\textbf{U}} &
  \multicolumn{1}{l|}{\textbf{HM}} &
  \textbf{AUC} &
  \multicolumn{1}{l|}{\textbf{S}} &
  \multicolumn{1}{l|}{\textbf{U}} &
  \multicolumn{1}{l|}{\textbf{HM}} &
  \textbf{AUC} &
  \multicolumn{1}{l|}{\textbf{S}} &
  \multicolumn{1}{l|}{\textbf{U}} &
  \multicolumn{1}{l|}{\textbf{HM}} &
  \textbf{AUC} \\ \hline
AoP~\cite{nagarajan2018attributesasoperators} &
  \multicolumn{1}{l|}{14.3} &
  \multicolumn{1}{l|}{17.4} &
  \multicolumn{1}{l|}{9.9} &
  1.6 &
  \multicolumn{1}{l|}{59.8} &
  \multicolumn{1}{l|}{54.2} &
  \multicolumn{1}{l|}{40.8} &
  25.9 &
  \multicolumn{1}{l|}{17.0} &
  \multicolumn{1}{l|}{5.6} &
  \multicolumn{1}{l|}{5.9} &
  0.7 \\ \hline
LE+~\cite{misra2017le+} &
  \multicolumn{1}{l|}{15.0} &
  \multicolumn{1}{l|}{20.1} &
  \multicolumn{1}{l|}{10.7} &
  2.0 &
  \multicolumn{1}{l|}{53.0} &
  \multicolumn{1}{l|}{61.9} &
  \multicolumn{1}{l|}{41.0} &
  25.7 &
  \multicolumn{1}{l|}{18.1} &
  \multicolumn{1}{l|}{5.6} &
  \multicolumn{1}{l|}{6.1} &
  0.8 \\ \hline
TMN~\cite{purushwalkam2019tmn} &
  \multicolumn{1}{l|}{20.2} &
  \multicolumn{1}{l|}{20.1} &
  \multicolumn{1}{l|}{13.0} &
  2.9 &
  \multicolumn{1}{l|}{58.7} &
  \multicolumn{1}{l|}{60.0} &
  \multicolumn{1}{l|}{45.0} &
  29.3 &
  \multicolumn{1}{l|}{23.1} &
  \multicolumn{1}{l|}{6.5} &
  \multicolumn{1}{l|}{7.5} &
  1.1 \\ \hline
SymNet~\cite{li2020symnet} &
  \multicolumn{1}{l|}{24.2} &
  \multicolumn{1}{l|}{25.2} &
  \multicolumn{1}{l|}{16.1} &
  3.0 &
  \multicolumn{1}{l|}{49.8} &
  \multicolumn{1}{l|}{57.4} &
  \multicolumn{1}{l|}{40.4} &
  23.4 &
  \multicolumn{1}{l|}{26.8} &
  \multicolumn{1}{l|}{10.3} &
  \multicolumn{1}{l|}{11.0} &
  2.1 \\ \hline
Compcos~\cite{mancini2021compcos} &
  \multicolumn{1}{l|}{25.3} &
  \multicolumn{1}{l|}{24.6} &
  \multicolumn{1}{l|}{16.4} &
  4.5 &
  \multicolumn{1}{l|}{59.8} &
  \multicolumn{1}{l|}{62.5} &
  \multicolumn{1}{l|}{43.1} &
  28.1 &
  \multicolumn{1}{l|}{28.1} &
  \multicolumn{1}{l|}{10.1} &
  \multicolumn{1}{l|}{12.4} &
  2.3 \\ \hline
CGE$_{ff}$~\cite{naeem2021cge} &
  \multicolumn{1}{l|}{28.7} &
  \multicolumn{1}{l|}{25.3} &
  \multicolumn{1}{l|}{17.2} &
  5.1 &
  \multicolumn{1}{l|}{56.8} &
  \multicolumn{1}{l|}{63.6} &
  \multicolumn{1}{l|}{41.2} &
  26.4 &
  \multicolumn{1}{l|}{28.1} &
  \multicolumn{1}{l|}{10.1} &
  \multicolumn{1}{l|}{11.4} &
  2.3 \\ \hline
Co-CGE$_{ff}$~\cite{mancini2022gcnopen} &
  \multicolumn{1}{l|}{27.8} &
  \multicolumn{1}{l|}{25.2} &
  \multicolumn{1}{l|}{17.5} &
  5.1 &
  \multicolumn{1}{l|}{58.2} &
  \multicolumn{1}{l|}{63.3} &
  \multicolumn{1}{l|}{44.1} &
  29.1 &
  \multicolumn{1}{l|}{29.3} &
  \multicolumn{1}{l|}{11.9} &
  \multicolumn{1}{l|}{12.7} &
  2.8 \\ \hline
CGE~\cite{naeem2021cge} &
  \multicolumn{1}{l|}{\textbf{32.8}} &
  \multicolumn{1}{l|}{28.0} &
  \multicolumn{1}{l|}{\textbf{21.4}} &
  6.5 &
  \multicolumn{1}{l|}{\textbf{64.5}} &
  \multicolumn{1}{l|}{\textbf{71.5}} &
  \multicolumn{1}{l|}{\textbf{60.5}} &
  33.5 &
  \multicolumn{1}{l|}{\textbf{33.5}} &
  \multicolumn{1}{l|}{15.5} &
  \multicolumn{1}{l|}{{16.0}} &
  4.2 \\ \hline
Co-CGE~\cite{naeem2021cge} &
  \multicolumn{1}{l|}{32.1} &
  \multicolumn{1}{l|}{\textbf{28.3}} &
  \multicolumn{1}{l|}{20.0} &
  6.6 &
  \multicolumn{1}{l|}{62.3} &
  \multicolumn{1}{l|}{66.3} &
  \multicolumn{1}{l|}{48.1} &
  33.9 &
  \multicolumn{1}{l|}{33.3} &
  \multicolumn{1}{l|}{14.9} &
  \multicolumn{1}{l|}{15.5} &
  4.1 \\ \hline
SCEN~\cite{li2022scen} &
  \multicolumn{1}{l|}{29.9} &
  \multicolumn{1}{l|}{25.2} &
  \multicolumn{1}{l|}{18.4} &
  5.3 &
  \multicolumn{1}{l|}{63.5} &
  \multicolumn{1}{l|}{66.3} &
  \multicolumn{1}{l|}{47.8} &
  32.0 &
  \multicolumn{1}{l|}{28.9} &
  \multicolumn{1}{l|}{12.1} &
  \multicolumn{1}{l|}{12.4} &
  2.9 \\ \hline
CAPE$_{ff}$ (Ours) &
  \multicolumn{1}{l|}{30.5} &
  \multicolumn{1}{l|}{26.2} &
  \multicolumn{1}{l|}{19.1} &
  5.8 &
  \multicolumn{1}{l|}{60.4} &
  \multicolumn{1}{l|}{67.4} &
  \multicolumn{1}{l|}{45.5} &
  31.3 &
  \multicolumn{1}{l|}{32.9} &
  \multicolumn{1}{l|}{15.6} &
  \multicolumn{1}{l|}{16.3} &
  4.2 \\ \hline
\textbf{CAPE} (Ours) &
  \multicolumn{1}{l|}{32.1} &
  \multicolumn{1}{l|}{28.0} &
  \multicolumn{1}{l|}{20.4} &
  \textbf{6.7} &
  \multicolumn{1}{l|}{62.3} &
  \multicolumn{1}{l|}{68.5} &
  \multicolumn{1}{l|}{49.5} &
  \textbf{35.2} &
  \multicolumn{1}{l|}{33.0} &
  \multicolumn{1}{l|}{\textbf{16.4}} &
  \multicolumn{1}{l|}{\textbf{16.3}} &
  \textbf{4.6} \\ \hline
\end{tabular}
\caption{Results on MIT-States, UT-Zappos and C-GQA. We report best seen (S) accuracy, best unseen (U) accuracy, best harmonic mean (HM), and area under the curve (AUC) on the compositions. ff denotes frozen feature extractor. Our model outperforms prior methodologies on AUC in all the datasets. }
\label{tbl:all_results}
\end{table*}

\subsection{Results}
Our approach outperforms all methodologies in AUC with especially significant improvement in CGQA~\cite{naeem2021cge}, the most challenging dataset. We achieve SOTA AUC of 4.6\% in CGQA dataset as compared to the last best result of CGE~\cite{naeem2021cge} of 4.2\%. We outperform unseen accuracy (U) and Harmonic Mean (HM). We set a new state-of-the-art of 16.3\% in Harmonic Mean and 16\% in unseen accuracy. In seen accuracy, we are comparable with the previous state-of-the-art CGE~\cite{naeem2021cge} by achieving 33.0\%. This impressive performance demonstrates our method's scalability to a large compositional space.\\
In the MIT-States dataset, we set a new state of the art in AUC by achieving 6.7\%. In Seen and Unseen accuracy, state of art results is set by \cite{naeem2021cge} by achieving 32.8\% and 28.3\% respectively. We are comparable to the previous state-of-the-art in seen and unseen accuracy by achieving 32.1\% and 28.0\%, respectively. MIT-States~\cite{isola2015mitstates} dataset has considerable label noise that can affect the interdependency structure discovery. However, our comparable results with the previous state-of-the-art show robustness to this label noise. \\
UT-Zappos is the smallest dataset and contains compositions that can not be differentiated visually~\cite{naeem2021cge}. We set a new state of the art by achieving 35.2\% AUC. We achieve 68.5\% unseen accuracy and 62.3\% seen accuracy, that is comparable to the previous state of the art.\\
We observe that we outperform all baseline methods consistently on large search spaces like CGQA~\cite{naeem2021cge}. This is because while previous algorithms only use a basic notion of compositional interdependency structure, we focus on a more complex view by integrating the discovery of the interdependency structure in our approach. This leads to the discovery of beneficial connections between different related compositions. \\
\myparagraph{Visualizing Propagation routes.} Table \ref{tbl:mostlikelyleastlikely} shows the top 5 and bottom 5 activations in the attention map from Multi-Head attention. 
% The attention map is taken before applying softmax. 
The listed compositions have the top five and bottom five actions for a given query. We conduct this analysis on CGQA~\cite{naeem2021cge} and MIT-States~\cite{isola2015mitstates} dataset. We report interesting compositions from all heads of Multi-Head attention. Results in Table \ref{tbl:mostlikelyleastlikely} confirm our hypothesis that a complex interdependency structure exists between compositions that do not share primitives. ``Cracked Mud" is related with ``Cracked Window" and ``Shattered Window". Shattered Window composition may visually contain a broken mirror with cracks visually similar to the state ``Cracked". Learning how ``Shattered" looks also updates how ``Cracked" looks. We observe the same with the composition ``Sliced Salmon" related to ``Pureed Seafood". While ``Sliced Salmon" and ``Pureed Seafood" do not share any primitive, they contain visual information as they come from a similar family of dishes containing seafood.
% the presence of ``Sliced Salmon" can guide the presence of ``Pureed Seafood".
Furthermore, ``Blue Mug" is related to ``Cluttered Kitchen" and ``Cluttered Desk". An image of ``Cluttered Kitchen" might also contain a blue mug. Learning how ``Blue Mug" looks will also help determine a ``Cluttered Kitchen" or ``Cluttered Desk" due to the propagation of knowledge. We also observe that the bottom 5 activation always contain completely irrelevant compositions. Such as, ``Red Floor" has the least activation for ``Green Salad", a food category. The same is observed with ``Winter Picture", which has least activation for ``Yellow Desk". On the other hand, ``Winter Picture" has high activations for ``Leafless", ``Barren", ``Forested", and ``Tree", and these objects can be found in an image of a ``Winter Picture".
Furthermore, object ``Redwood" in ``Weathered Redwood" have high activations for object ``Log" that are both representation of wood. Similarly, states ``Weathered", ``Broken", ``Splintered" are related with each other such that all of them represent a damaged or worn out state of an object. 
We also observe that our approach is able to find propagation routes for compositions with shared primitives such as, ``Yellow Wall" have highest activation for ``Yellow Chair", ``Dry Pond" have high activations for ``Dry Bush" and ``Dry Forest" and ``Weathered Redwood" have high activation for ``Weathered Log". This shows that our approach can find simple propagation routes that share primitives and also complex propagation routes share some property of state or object and not expressed in primitives. \\
\textbf{Impact of Feature Representations. } Consistent with previous works, we experimented with the frozen backbone in our network. CAPE$_{ff}$ represents our approach with frozen backbone in Table \ref{tbl:all_results}. We outperform all previous methods with a frozen backbone, including a recent approach SCEN~\cite{li2022scen} that is specially developed for only frozen features. In CGQA~\cite{naeem2021cge}, we match state of the art from CGE~\cite{naeem2021cge} in AUC and outperform in harmonic mean and unseen accuracy. In UT-Zappos~\cite{yu2014utzap1,yu2017utzap2}, CAPE$_{ff}$ outperforms in unseen accuracy over previous approaches with frozen backbone. While in the MIT-States dataset, CAPE$_{ff}$ outperforms previous approaches with the frozen backbone in all metrics.

\begin{table*}[t]
\centering
\resizebox{\textwidth}{!}{
\begin{tabular}{|l|ll|}
\hline
 &
  \multicolumn{2}{l|}{\textbf{MIT-States}} \\ \hline
\textbf{Query} &
  \multicolumn{1}{l|}{\textbf{Top 5}} &
  \textbf{Bottom 5} \\ \hline
\textit{Cracked Mud} &
  \multicolumn{1}{l|}{\begin{tabular}[c]{@{}l@{}}Cracked Window, Shattered Window, Cracked Door, \\ Broken Window, Cracked Mirror\end{tabular}} &
  \begin{tabular}[c]{@{}l@{}}Fresh Bread, Fresh Butter, Fresh Cheese, \\ Fresh Meat, Fresh Orange\end{tabular} \\ \hline
\textit{Weathered Redwood} &
  \multicolumn{1}{l|}{\begin{tabular}[c]{@{}l@{}}Broken Log, Splintered Log, Peeled Log, \\ Burnt Log, Weathered Log\end{tabular}} &
  \begin{tabular}[c]{@{}l@{}}Old Bus, Old City, Old Car, \\ Old Truck, Old Street\end{tabular} \\ \hline
\textit{Rusty Bridge} &
  \multicolumn{1}{l|}{\begin{tabular}[c]{@{}l@{}}Old Log, Old Library, Old Boat, \\ Old Computer, Old Bear\end{tabular}} &
  \begin{tabular}[c]{@{}l@{}}Broken Column, Broken City, Broken Lightbulb, \\ Broken Jewelry, Broken Necklace\end{tabular} \\ \hline
\textit{Dry Pond} &
  \multicolumn{1}{l|}{\begin{tabular}[c]{@{}l@{}}Dry Bush, Wet Bush, Damp Bush, \\ Barren Bush, Dry Forest\end{tabular}} &
  \begin{tabular}[c]{@{}l@{}}Eroded Shore, Broken Shore, Weathered Shore, \\ Verdant Shore, Mossy Shore\end{tabular} \\ \hline
\textit{Sliced Salmon} &
  \multicolumn{1}{l|}{\begin{tabular}[c]{@{}l@{}}Pureed Seafood, Pureed Fish, Pureed Salmon, \\ Diced Seafood, Cooked Seafood\end{tabular}} &
  \begin{tabular}[c]{@{}l@{}}Ruffled Bed, Wide Blade, Draped Bed, \\ Ruffled Shower, Ruffled Leaf\end{tabular} \\ \hline
 &
  \multicolumn{1}{l|}{\textbf{CGQA}} &
   \\ \hline
\textbf{Query} &
  \multicolumn{1}{l|}{\textbf{Top 5}} &
  \textbf{Bottom 5} \\ \hline
\textit{Red Floor} &
  \multicolumn{1}{l|}{\begin{tabular}[c]{@{}l@{}}Carpeted Floor, Textured Floor, Cracked Floor, \\ Painted Floor, Sripped Floor\end{tabular}} &
  \begin{tabular}[c]{@{}l@{}}Green Salad, Green Brocoli, Green Cabbage, \\ Green Asparagus, Green Apple\end{tabular} \\ \hline
\textit{Winter Picture} &
  \multicolumn{1}{l|}{\begin{tabular}[c]{@{}l@{}}Overgrown Tree, Forested Tree, Leafless Tree, \\ Overgrown Weeds, Barren Tree\end{tabular}} &
  \begin{tabular}[c]{@{}l@{}}Yellow Desk, Comfortable Chair, Yellow Chair, \\ Red Desk, Plaid Chair\end{tabular} \\ \hline
\textit{Large Cooler} &
  \multicolumn{1}{l|}{\begin{tabular}[c]{@{}l@{}}Large Snow, Large Crust, Large Mountain, \\ Large Omelette, Large Barier\end{tabular}} &
  \begin{tabular}[c]{@{}l@{}}Full Hangar, Full Floor, Full Ground, \\ Open Door, Transparent Door\end{tabular} \\ \hline
\textit{Blue Mug} &
  \multicolumn{1}{l|}{\begin{tabular}[c]{@{}l@{}}Cluttered Kitchen, Cluttered Desk, Cluttered Shelf, \\ Cluttered Office, Cluttered Counter\end{tabular}} &
  \begin{tabular}[c]{@{}l@{}}Leafless Bush, Leaflless Tree, Leafless Branch, \\ Bushy Bush, Bamboo Bush\end{tabular} \\ \hline
\textit{Yellow Wall} &
  \multicolumn{1}{l|}{\begin{tabular}[c]{@{}l@{}}Yellow Chair, Colorful Chair, Red Chair, \\ Purple Chair, Striped Chair\end{tabular}} &
  \begin{tabular}[c]{@{}l@{}}Bare Tree, Huge Tree, Overgrown Tree, \\ High Tree, Tall Tree\end{tabular} \\ \hline
\end{tabular}
}

\caption{The table shows the Top 5 and bottom 5 pairs for the query pairs given in the first column taken from MIT-States and CGQA datasets. Top 5 and bottom 5 pairs are selected from the attention matrix before softmax. We observe that query compositions have highest activations for similar compositions shown in column ``Top 5" and least activations for different compositions shown in column ``Bottom 5". We also observe that our approach can find diverse propagation routes. }
\label{tbl:mostlikelyleastlikely}
\end{table*}

\begin{figure*}[h]
\centering
\includegraphics[width=\textwidth]{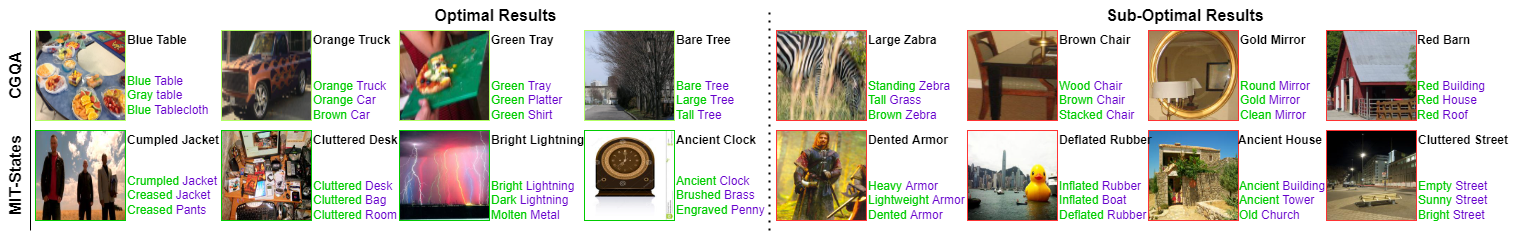}
\caption{Shows qualitative results of our approach on MIT-States and CGQA dataset. First four images in each row from the left show positive results where our approach predicted correct composition. Last four image show sub optimal results where our approach did not predict correct composition. First row contains results from CGQA dataset and second row contains results from MIT-States dataset.}
\label{fig:qualitative_images}
\end{figure*}
\subsection{Qualitative Results}
In this section, we will discuss our qualitative results in detail. Figure \ref{fig:qualitative_images} shows qualitative results of our approach on MIT-States~\cite{isola2015mitstates} and CGQA~\cite{naeem2021cge} dataset. We observe that our approach can predict compositions of images containing noise and other valid compositions. Composition ``Crumpled Jacket" contains objects clouds and grass. Likewise, the composition ``Bare Tree" also contain objects road and a car. The composition ``Cluttered Desk" contains objects table and frame. The composition "Green Tray" contains object ``Pizza" as well. \\
We also show sub-optimal results where our approach did not predict compositions correctly. We observe that in multiple sub-optimal results, the correct composition is present in the top 3 predictions. Sub-optimal results that contain correct predictions in the top 3 results are ``Dented Armor", ``Deflated Rubber", ``Brown Chair", and ``Gold Mirror". We also observe label noise in MIT-States results, such as ``Deflated Rubber" is a duck floating because it is ``Inflated", as predicted by our approach. Furthermore, ``Dented Armor" does not show any dents in it, and our approach predicts it as ``Heavy Armor" which is more close to the visual representation of the image.
We also observe that our sub-optimal predictions are not necessarily incorrect. For example, composition ``Gold Mirror" is a circular mirror, and our approach predicts it as "Round Mirror". Composition "Brown Chair" predicted as ``Wooden Chair" is a chair made of wood. Composition ``Large Zebra" predicted as ``Standing Zebra" shows an image of ``Zebra" while standing. There are multiple valid compositions for a given image. We come to the same conclusion as \cite{naeem2021cge} that problem of CZSL should be considered a multi-label problem.

\section{Ablation}
In this section, we ablate over different configurations of the attention mechanism of CAPE. Results of ablation for harmonic mean and AUC are shown in Table \ref{tbl:ablation}. CAPE in Table \ref{tbl:ablation} represents our original approach proposed in section \ref{sec:approach}. We conduct our ablation on MIT-States~\cite{isola2015mitstates} and CGQA~\cite{naeem2021cge} datasets. Results are reported on validation sets from both datasets. Our original approach as proposed in section \ref{sec:approach} achieves 8.2\% AUC and 23.2\% Harmonic mean on MIT-States~\cite{isola2015mitstates} dataset. It achieves 6.1\% AUC and 19.5\% harmonic mean on CGQA~\cite{naeem2021cge} dataset. The architecture of the CAPE-Propagator does not change during these experiments. In all experiments, number of heads in Multi-Head Attention are kept constant at 6. We keep all hyperparameters constant across all experiments.\\
\textbf{Self Attention on States, Objects and Compositions (CAPE$_{self}$).} This configuration is represented by CAPE$_{self}$ in Table \ref{tbl:ablation}. We create one tensor $\hat{\mathcal{Y}}_{self}$ where $|\hat{\mathcal{Y}}_{self}|= |\mathcal{S}|+ |\mathcal{O}| + |\mathcal{Y}_s|$ and containins word embeddings of states, objects and compositions. During testing, we append unseen compositions to $\hat{\mathcal{Y}_{self}}$ to get $|\hat{\mathcal{Y}}_{self}|= |\mathcal{S}|+|\mathcal{O}| + |\mathcal{Y}_s| + |\mathcal{Y}_n|$. Compositions are calculated as mean of their state and object word embeddings. CAPE-Propagator projects $\hat{\mathcal{Y}_{self}}$ to shared semantic space to get $\mathcal{Y}_{F}$. Final composition scores are calculated between image embedding and compositional nodes in $\mathcal{Y}_{F}$ by using compatibility function shown in Eq. \ref{eq:compatibiltiy}. During training CAPE-Propagator learns propagation routes by exploiting self-attention mechanism as explained in section \ref{sec:approach}. CAPE$_{self}$ configuration is very similar to the configuration used in \cite{naeem2021cge} that also applies supervision on only compositional nodes. CAPE$_{self}$ achieves 8.1\% AUC and 23.0\% harmonic mean on the MIT-States dataset. It achieves 6.0\% AUC and 19.2\% harmonic mean on CGQA~\cite{naeem2021cge} dataset. We observe a performance loss as CAPE$_{self}$ lags in both datasets in HM and AUC. \\
\textbf{Cross Attention on Primitives and Self Attention on Compositions (CAPE$_{dual}$). } This configuration is represented by CAPE$_{dual}$ in Table \ref{tbl:ablation}. CAPE$_{dual}$ applies cross attention on states and objects and self-attention to their compositions. Firstly, we apply cross attention on word embeddings of state and objects. We use one Multi-Head attention to apply cross attention between states and objects to get $\hat{\mathcal{Y}_{states}}$. We use second Multi-Head attention to apply cross attention between objects and states to get $\hat{\mathcal{Y}_{objects}}$. Afterwards, we concatenate $\hat{\mathcal{Y}_{states}}$ and $\hat{\mathcal{Y}_{objects}}$ to get compositions $\hat{\mathcal{Y}_{dual}}$ where $|\hat{\mathcal{Y}_{dual}}|=|\mathcal{Y}_s|$ during training and $|\hat{\mathcal{Y}_{dual}}|=|\mathcal{Y}_s| + |\mathcal{Y}_n|$ during testing. We input $\hat{\mathcal{Y}_{dual}}$ into CAPE-Propagator to get $\mathcal{Y}_F$. CAPE-Propagator discovers propagation routes on $\hat{\mathcal{Y}_{dual}}$ by exploiting self-attention mechanism as explained in section \ref{sec:approach}. Final composition scores are calculated between image embeddings and $\mathcal{Y}_F$ by using compatibility function shown in Eq. \ref{eq:compatibiltiy}. CAPE$_{dual}$ achieves 8.2\% AUC and 23.2\% harmonic mean on MIT-States dataset~\cite{isola2015mitstates}. It achieves 6.0\% AUC and 19.5\% harmonic mean on CGQA~\cite{naeem2021cge} dataset. This configuration matches the performance of CAPE in MIT-States~\cite{isola2015mitstates} dataset but lags behind in AUC in CGQA~\cite{naeem2021cge} dataset. \\
\textbf{Multi-Layer Perceptron (MLP) as a replacement for CAPE-Propagator. } In this experiment, we used Multi-Layer perceptron instead of CAPE-Propagator or Multi-Head attention. The results are represented by heading ``MLP" in Table \ref{tbl:ablation}. MLP was configured to have same amount of parameters as CAPE-Propagator. MLP achieves 7.5\% AUC and 22.1\% harmonic mean on MIT-States~\cite{isola2015mitstates} dataset. It achieves 5.4\% AUC and 18.3\% harmonic mean on CGQA~\cite{naeem2021cge} dataset. Since MLP does not model interdependency structure, it lags in all datasets. \\
We observe that our original configuration, as proposed in section \ref{sec:approach} outperforms all configurations. Introducing additional embeddings or MHAs leads to poorer performance. CAPE is an effective approach that exploits self-attention to learn hidden interdependency structures between compositions caused by the primitives. Introducing new networks, like in the case of CAPE$_{dual}$ leads to an increase in the number of learnable parameters and a redundant cross-attention mechanism. On the other hand, primitives do not get supervision in CAPE$_{self}$ leading to poorer performance. 

\begin{table}[h]
\centering
\begin{tabular}{|l|ll|ll|}
\hline
           & \multicolumn{2}{l|}{\textbf{MIT-States}} & \multicolumn{2}{l|}{\textbf{CGQA}}       \\ \hline
           \textbf{Method} & \multicolumn{1}{l|}{\textbf{AUC}} &\textbf{HM}   & \multicolumn{1}{l|}{\textbf{AUC}} & \textbf{HM}   \\ \hline
CAPE$_{dual}$ & \multicolumn{1}{l|}{\textbf{8.2}} & \textbf{23.2} & \multicolumn{1}{l|}{6.0} & \textbf{19.5} \\ \hline
CAPE$_{self}$ & \multicolumn{1}{l|}{8.1} & 23.0 & \multicolumn{1}{l|}{6.0} & 19.2 \\ \hline
MLP        & \multicolumn{1}{l|}{7.5} & 22.1 & \multicolumn{1}{l|}{5.4} & 18.3 \\ \hline
\textbf{CAPE}       & \multicolumn{1}{l|}{\textbf{8.2}} & \textbf{23.2} & \multicolumn{1}{l|}{\textbf{6.1}} & \textbf{19.5} \\ \hline
\end{tabular}
\caption{\textit{Ablation} over different configurations of CAPE. We report highest achieved AUCs on MIT-States~\cite{isola2015mitstates} and CGQA~\cite{naeem2021cge} validation dataset. }
\label{tbl:ablation}
\end{table}

\section{Conclusion}
We propose a novel approach to the task of Compositional Zero-Shot Learning. We evaluated our approach on three benchmark datasets (CGQA~\cite{naeem2021cge}, MIT-States~\cite{isola2015mitstates}, UT-Zappos~\cite{yu2014utzap1,yu2017utzap2}) extensively. We argued that there is a complex interdependency structure between compositions that do not share any primitives. We exploited the attention mechanism to discover this interdependency structure and propagate it to unseen classes. Our qualitative analysis reaffirm our original hypothesis that there exists a complex interdependency structure between compositions. Our approach outperforms prior methodologies and shows improvement in all benchmark datasets. Our qualitative results demonstrate that there can be multiple valid predictions of one image, and the problem of CZSL should be considered a multi-label problem. We encourage future works to consider this aspect while building their methodologies. \\
\myparagraph{Acknowledgement.} This work was partially funded by the European project FLUENTLY (101058680) and partially by the
European project INFINITY (883293).

%-------------------------------------------------------------------------
\newpage
{\small
\bibliographystyle{ieee_fullname}
\bibliography{egbib}
}

\end{document}